\documentclass[conference,10pt,letter]{IEEEtran}
\def\FinalVersion{}
\newcounter{ExampleCounter}
\RequirePackage[l2tabu,orthodox,abort]{nag}
\RequirePackage[english]{babel}
\RequirePackage[utf8]{inputenc}
\RequirePackage[T1]{fontenc}
\usepackage{parskip}

\RequirePackage{comment}
\RequirePackage{url}
\RequirePackage{subfig}

\RequirePackage{graphicx} 
\RequirePackage{amsmath}
\RequirePackage{amssymb}
\RequirePackage{amsthm}    

\RequirePackage{bm}       
\RequirePackage{textcomp} 
\RequirePackage{acronym}
\RequirePackage{algorithm}
\RequirePackage{marginnote}

\RequirePackage[hang]{footmisc}
\RequirePackage[
    locale = UK,
    strict,
    forbid-literal-units,   
    parse-numbers = false,  
    parse-units = false,    
    sticky-per,             
    ]{siunitx}

\RequirePackage[colorinlistoftodos]{todonotes}
\let\marginnoteOLD\marginnote
\ifdefined\FinalVersion
\renewcommand{\marginnote}[1]{}
\else
\renewcommand{\marginnote}[1]{\marginnoteOLD{\small{#1}}}
\fi

\newcommand{\MyTitle}{%
    Conservative, Proportional and Optimistic Contextual Discounting in the Belief Functions Theory%
}

\newcommand{\MyKeywords}{%
    contextual discounting, 
    temporal discounting, 
    discounting,
    belief functions, 
    Dempster-Shafer theory,
    evidence theory,
    contextual fusion%
}

\newcommand{\V}[1]{%
    \bm{#1}
}
\newcommand*\circled[1]{%
    \mathchoice
    {
        \begin{tikzpicture}
        \node[draw,circle,inner sep=1pt] {%
            \footnotesize
            {\( #1 \)}
        };
        \end{tikzpicture}
    }
    {
        \begin{tikzpicture}
        \node[draw,circle,inner sep=1pt] {%
            {\( #1 \)}
        };
        \end{tikzpicture}
    }
    {
        \begin{tikzpicture}
        \node[draw,circle,inner sep=1pt] {%
            {\( #1 \)}
        };
        \end{tikzpicture}
    }
    {
        \begin{tikzpicture}
        \node[draw,circle,inner sep=1pt] {%
            \tiny
            {\( #1 \)}
        };
        \end{tikzpicture}
    }
}

\newcommand{\oppause}{\:}

\newcommand{\opdisj}{\oppause\circled{\cup}\oppause}

\newcommand{\bigop}[1]%
{%
    \mathop%
    {%
        \vphantom{\sum}%
        \mathchoice%
        {\vcenter{\hbox{\huge{$#1$}}}}%
        {\vcenter{\hbox{\Large{$#1$}}}}%
        {#1}%
        {#1}%
    }%
    \displaylimits%
}

\newenvironment{keyword}{\noindent\small\textbf{Keywords:}\enspace\ignorespaces}{}

\newcommand{\email}[1]{%
    \href{mailto:#1}{\texorpdfstring{\texttt{{#1}}}{#1}}%
}
\newcommand{\website}[1]{%
    \texttt{\url{#1}}%
}
\newcommand{\quotemarks}[1]{``#1''}

\newcommand{\Note}[1]{\emph{Note:}\enspace\ignorespaces#1}
\RequirePackage{hyperref}
\hypersetup{
    bookmarksnumbered=true,
    bookmarksopen=true,
    bookmarksopenlevel=2,
    colorlinks=true, 
    linkcolor=black,
    citecolor=black, 
    filecolor=black,
    urlcolor=black,
    pdfinfo={
        Title={\MyTitle},
        Keywords={\MyKeywords},
        Author={Marek Kurdej, V\'{e}ronique Cherfaoui},
    }
}
\acrodef{BBA}[bba]{basic belief assignment}
\acrodef{CRC}{conjunctive rule of combination}
\acrodef{DRC}{disjunctive rule of combination}
\acrodef{FOD}[fod]{frame of discernment}
\acrodef{FODs}[fods]{frames of discernment}
\acrodef{DST}{Dempster--Shafer theory}
\acrodef{TBM}{Transferable Belief Model}
\acrodef{TEF}{Temporal Evidential Filter}
\DeclareMathOperator{\bel}{bel}
\DeclareMathOperator{\mm}{m}

\DeclareMathOperator{\vv}{\nu}

\IEEEoverridecommandlockouts
\title{\MyTitle} 
\begin{document}
\author{%
    \IEEEauthorblockN{
        Marek Kurdej
    }
    \IEEEauthorblockA{
        \email{marek.kurdej@hds.utc.fr}
        \\
        \website{http://www.hds.utc.fr/~kurdejma}
        \\
        \href{http://www.utc.fr}{Universit\'{e} de Technologie de Compi\`{e}gne} -- CNRS,
        \\
        \href{http://www.hds.utc.fr}{Heudiasyc} UMR 7253,
        \\
        BP 20529,
        \MakeUppercase{60205 Compi\`{e}gne cedex},
        France
    }
    \and%
    \IEEEauthorblockN{
        V\'{e}ronique Cherfaoui
    }
    \IEEEauthorblockA{
        \email{veronique.cherfaoui@hds.utc.fr}
        \\
        \website{http://www.hds.utc.fr/~vberge}
        \\
        \href{http://www.utc.fr}{Universit\'{e} de Technologie de Compi\`{e}gne} -- CNRS,
        \\
        \href{http://www.hds.utc.fr}{Heudiasyc} UMR 7253,
        \\
        BP 20529,
        \MakeUppercase{60205 Compi\`{e}gne cedex},
        France
    }
}
\date{}
\maketitle

\begin{abstract}
    Information discounting plays an important role in the theory of belief functions and, generally, in information fusion.
    Nevertheless, neither classical uniform discounting nor contextual cannot model certain use cases, notably temporal discounting.
    In this article, new contextual discounting schemes, conservative, proportional and optimistic, are proposed.
    Some properties of these discounting operations are examined.
    Classical discounting is shown to be a special case of these schemes.
    Two motivating cases are discussed: modelling of source reliability and application to temporal discounting.
\end{abstract}

\setcounter{tocdepth}{4}

\section{Introduction}
\label{sec:introduction}

In many problems of information fusion, there is a need to allow for the reliability of a source \cite{Smets2000a}.
The meta-knowledge about the reliability can be only source-dependent, but it can as well vary for different types of evidence.
While the first case is easily handled by classical discounting operation \cite{Shafer1976}, the second one is more complex and existing solutions do not meet all possible use cases \cite{Mercier2006,Mercier2010}.
In this article, we address this problem in the context of the theory of belief functions, also known as Dempster--Shafer theory \cite{Dempster1968,Shafer1976} by proposing three schemes for contextual discounting: conservative, proportional and optimistic.

The domain of information fusion concerns in great measure the combination of sensor data arriving successively with the passage of time.
Past information is often useful and should not be discarded.
However, one cannot disregard the fact that the information may worth less and less over time.
In order to handle this variation in subjective value of a piece of information, we apply proposed discounting operations to temporal discounting.

\ifdefined\FinalVersion\else
\subsection*{Related work}
\fi
\label{related-work}

The closest work and the starting point for this article has been realised by Mercier et~al. who presented the original idea of contextual discounting \cite{Mercier2005, Mercier2006}.
This research has been further developed and generalised contextual discounting and reinforcement have been described as examples of correction mechanisms for belief functions \cite{Mercier2010, Mercier2012}.
Pichon et~al. devoted some research to the subject of information correction schemes by proposing a strategy taking into account the source's relevance and truthfulness \cite{Pichon2012}.
Other mechanisms of data revision have been studied in the context of the evidence theory.
A review of existing revision rules can be found in \cite{Ma2011}, along with an extension of one of them able to cope with inconsistency between prior and input information.

\ifdefined\FinalVersion\else
\subsection*{Article organisation}
\fi
\label{organisation}

The rest of this paper is organised as follows.
The existing concepts of discounting in the theory of belief functions will first be recalled in Section~\ref{sec:fundamentals}.
Next, Section~\ref{sec:discounting} will present the details of the proposed schemes.
Rules' behaviour and their properties will be described in Section~\ref{sec:behaviour}, while some simple examples will be given in Section~\ref{sec:examples}.
A case study about the application of the proposed method to temporal discounting will be the subject of Section~\ref{sec:case-study}.
We will conclude the paper and outline the perspectives for future research in Section~\ref{sec:conclusion}.

\section{Belief functions theory}
\label{sec:fundamentals}

\subsection{Fundamentals}
\label{fundamentals}

The information obtained from source $S$ concerning the actual value taken by variable $x$ is quantitatively described by \acf{BBA} \( \mm^{\Omega}_{S} \).
Variable $x$ takes values in a finite set \( \Omega = \{ \omega_{1}, \ldots, \omega_{K} \} \) which is called \acf{FOD}.
\( \mm^{\Omega}_{S} \) is defined as a function from $2^{\Omega}$ to interval $[0, 1]$ satisfying the condition:
\begin{align}
    \label{eq:mass-function}
    \sum\limits_{A \subseteq \Omega} \mm^{\Omega}_{S}(A) &= 1
\end{align}
The notation \( \mm^{\Omega}_{S} \) will be further simplified to \( \mm^{\Omega} \) or \( \mm \) when no ambiguity is possible.
Total ignorance about the variable $x$ is represented by a \emph{vacuous \ac{BBA}} for which \( \mm(\Omega) = 1 \).
Additionally, a mass function satisfying \( \mm(\emptyset) = 0 \) will be called \emph{normal} or \emph{regular}, whereas one not fulfilling this condition --- \emph{subnormal}.

In following sections, the \ac{DRC} will be used.
\ac{DRC} may be used to combine two distinct pieces of evidence $\mm_1$, $\mm_2$ under the assumption that at least one of the two information sources is reliable \cite{Smets2008}.
\ac{DRC} is defined by:
\begin{align}
    \label{eq:drc}
    (\mm_1 \opdisj \mm_2)(A)
    &= \sum\limits_{B \cup C = A} \mm_1(B) \mm_2(C)
    & \forall A \subseteq \Omega
\end{align}

A \acl{BBA} can be expressed not only by mass function $\mm$, but there are equivalent functions representing the same information.
One of them is belief function \( \bel : 2^{\Omega} \rightarrow [0, 1] \) which in the \ac{TBM} \cite{Smets1994tbm} take the form of:
\begin{align}
    \label{eq:mass-to-belief}
    \bel(A)
    &= \sum_{\emptyset \neq B \subseteq A} \mm(B)
    & \forall A \subseteq \Omega
\end{align}
%

\subsection{Classical discounting}
\label{classical-discounting}

The most commonly used form of discounting operation given discount factor $\alpha$ has been proposed by Shafer in \cite[pp.~251--255]{Shafer1976} and will be subsequently called \emph{classical discounting}:
\begin{align}
    \label{eq:classical-discounting}
    {}^{\alpha} \bel(A)
    &= (1 - \alpha) \bel(A)
    & \forall A \subsetneq \Omega
\end{align}
which can be expressed equivalently using mass functions as:
\begin{align}
    \label{eq:}
    {}^{\alpha} \mm(A)
    &= (1 - \alpha) \mm(A)
    & \forall A \subsetneq \Omega
    \\
    {}^{\alpha} \mm(\Omega)
    &= (1 - \alpha) \mm(\Omega) + \alpha
\end{align}

\subsection{Contextual discounting}
\label{contextual-discounting}

\emph{Contextual discounting}, an extension of classical discounting taking into account reliabilities varying between classes has been proposed by Mercier et.~al. \cite{Mercier2005} and developed in \cite{Mercier2006,Mercier2010,Mercier2012}.
This operation uses vector $\V{\alpha}$ of discount factors $\alpha_\theta$ attributed to elements $\theta$ of partition $\Theta$ of the \acl{FOD} $\Omega$, i.e.:
\begin{align}
    \label{eq:partition}
    \Theta &\subseteq 2^{\Omega}
    \\
    \Omega
    &= \bigcup_{\theta \in \Theta} \theta
    \\
    \forall \theta_{i}, \theta_{j} \in \Theta, i \neq j
    &: \theta_{i} \cap \theta_{j}
    = \emptyset
\end{align}
Contextual discounting \( {}^{\V{\alpha}}_{\Theta} \mm \) of a \ac{BBA} $\mm$ is equal to:
\begin{align}
    \label{eq:contextual-discounting}
    {}^{\V{\alpha}}_{\Theta} \mm
    &= \mm \opdisj \mm_{\Theta}
    \\
    \mm_{\Theta}
    &= \mm_{1} \opdisj \mm_{2} \opdisj \ldots \opdisj m_{L}
\end{align}
where each \( \mm_{l} \), \( l \in \{ 1,\ \ldots,\ L \} \), is defined by:
\begin{align}
    \label{eq:contextual-discounting-functions}
    \mm_{l} &= 
    \begin{cases}
        1 - \alpha_{l} & \text{ if } A = \emptyset
        \\
        \alpha_{l} & \text{ if } A = \theta_{l}
        \\
        0 & \text{ otherwise }
    \end{cases}
\end{align}
One of the inconveniences of this method is the fact that reliability factors are attributed to a partition of the frame of discernment, which excludes cases where reliability is known for intersecting subsets of $\Omega$.

\subsection{Generalised contextual discounting}
\label{generalised-contextual-discounting}

The aforementioned problem has been addressed in \cite{Mercier2010,Mercier2012} where \emph{generalised contextual discounting} is proposed as a correction mechanism.
Again, vector $\V{\alpha}$ of discount factors is used, but here, they can be defined also for intersecting sets.
The method employs canonical disjunctive decomposition of a subnormal \ac{BBA} introduced by Denœux \cite{Denoeux2008}.
The idea is to discount disjunctive weights $\vv$ of such a decomposition of \ac{BBA}~$\mm$:
\begin{align}
    \label{eq:generalised-contextual-discounting}
    {}^{\V{\alpha} \cup}_{\Theta} \mm
    &= \bigop{\opdisj}_{A \supset \emptyset} \V{A}_{(1 - \alpha_{A}) \cdot \vv(A)}
\end{align}
where $\V{A}_{\vv(A)}$ is a negative generalised simple \ac{BBA} (NGSBBA) \cite{Denoeux2008} defined from $2^{\Omega}$ to $\mathcal{R}$ by:
\begin{align}
    \label{eq:disjunctive-weight-function}
    \V{A}_{\vv(A)} :\ 
        \emptyset & \mapsto \vv(A)
        \\
        A & \mapsto 1 - \vv(A) & 
        \\
        B & \mapsto 0 & \forall B \in 2^{\Omega} \backslash \{ \emptyset, A \}
\end{align}

\section{Conservative, optimistic and proportional discounting}
\label{sec:discounting}

As a departure point for the design of an operation of discounting, a few hypotheses have been set.
First and foremost, information source $S$ is supposed to excessively encourage set of solutions $A$ and, therefore, $\mm(A)$ should be discounted by factor $\alpha_{A}$ corresponding to $A$.
The behaviour of the new discounting operation should be close to the behaviour of classical discounting.
Mass of conflict $\mm(\emptyset)$ shall get discounted and new schemes should generalise the classical one.
Moreover, setting a non-zero discount factor for set $\theta$ should entail the discounting of mass attributed to $\theta$, whereas masses of sets having no elements in common with $\theta$ should rest unchanged\footnote{Except for frame of discernment $\Omega$, since masses are transferred to this set.}.
Such a behaviour is opposite to contextual discounting proposed by Mercier et~al. \cite{Mercier2005} that retains mass attributed to $\theta$ and discounts \emph{other} sets, which we judge counter-intuitive especially in case of many classes, but well-justified and conform to the proposed interpretation (see \cite[Example~2]{Mercier2006}).
Finally, we postulate that discounted mass of set $\theta$ should be transferred to $\Omega$ and not to its other superset being a proper subset of $\Omega$, since doing so would imply additional knowledge about the state of the represented entity%
.

\subsection{Notation}
\label{notation}

In the following sections, we will stick to similar notation as in Section~\ref{sec:fundamentals}.
In order to distinguish proposed discounting operations between them and to avoid any confusion with existing schemes,
\( {}_{c, \Theta}^{\V{\alpha}} \mm \) will denote conservative discounting of a \ac{BBA} $\mm$ using discount rate vector $\V{\alpha}$ defined for all elements of \( \Theta \subseteq 2^{\Omega} \).
Similarly, \( {}_{p, \Theta}^{\V{\alpha}} \mm \) will represent proportional discounting and \( {}_{o, \Theta}^{\V{\alpha}} \mm \) --- optimistic discounting.

When the set of classes $\Theta$ for which discount factors are defined is obvious or unimportant, notation \( {}_{c}^{\V{\alpha}} \mm \) will be equivalent to \( {}_{c, \Theta}^{\V{\alpha}} \mm \).
Analogical convention will be used for other types of discounting.
Equally, we simplify the notation by omitting the type of discounting ($c$ for conservative, $p$ for proportional or $o$ for optimistic) if an equation is valid for all types.
Finally, set \( \{ \omega_{1}, \omega_{2} \} \) will be denoted \( \omega_{1} \omega_{2} \) and
\( \alpha_{\theta} \) will refer to the discount rate defined for set $\theta$, given that \( \theta \in \Theta \), \( \Theta \subseteq 2^{\Omega} \).

\subsection{Conservative discounting}
\label{conservative-discounting}

Conservative discounting presents a pessimistic approach to the discounting.
As stated before, the attribution of $\mm(A)$ by source $S$ is excessive and this mass should be discounted by $\alpha_{A}$.
Let us suppose now that some meta-knowledge states additionally that the affectation of masses to supersets of $A$ by source $S$ is highly dependent on class $A$.
Bearing in mind the above statement, the mass attributed to $AB$ should be discounted in the same manner as $\mm(A)$.
Therefore, in conservative discounting, set $\theta$, the empty set and all sets having at least one element in common with $\theta$ are discounted by the same factor $\alpha_{\theta}$.

Generalising this behaviour to any \( \Theta \subseteq 2^{\Omega} \), one obtains:
\begin{align}
    \label{eq:conservative-emptyset}
    {}^{\V{\alpha}}_{c} \mm(\emptyset)
        &= \mm(\emptyset) \cdot
        \prod\limits_{\theta \in \Theta}
            1 - \alpha_{\theta}
    \\
    \label{eq:conservative-set}
    {}^{\V{\alpha}}_{c} \mm(A)
        &= \mm(A) \cdot
        \prod\limits_{ \substack{\theta \in \Theta \\ A \cap \theta \neq \emptyset} }
            1 - \alpha_{\theta}
        \qquad \forall A \subsetneq \Omega, A \neq \emptyset
    \\
    \label{eq:conservative-omega}
    {}^{\V{\alpha}}_{c} \mm(\Omega)
        &= \mm(\Omega) \cdot
        \prod\limits_{ \theta \in \Theta }
            1 - \alpha_{\theta}
            \\&\nonumber
        + \mm(\emptyset) \cdot
            \prod\limits_{\theta \in \Theta}
                \alpha_{\theta}
            \\&\nonumber
        + \sum\limits_{ A \subseteq \Omega }
            \left[
                \mm(A) \cdot
                \prod\limits_{ \substack{\theta \in \Theta \\ A \cap \theta \neq \emptyset} }
                    \alpha_{\theta}
            \right]
\end{align}
One remarks that the most discounted mass is $\mm(\emptyset)$ which is affected by all discount rates.

\subsection{Optimistic discounting}
\label{optimistic-discounting}

Optimistic discounting is based on a hypothesis opposite to the one made in conservative discounting.
This time, the meta-information about source $S$ asserts that masses of supersets of $A$ are affected independently of class $A$.
These masses shall \emph{not} be discounted by $\alpha_A$.
On the other hand, all subsets of $A$ will be affected in the same way as $A$.

This type of discounting can be expressed for any \( \Theta \subseteq 2^{\Omega} \) by:
\begin{align}
    \label{eq:optimistic-emptyset}
    {}^{\V{\alpha}}_{o} \mm(\emptyset)
    &= \mm(\emptyset) \cdot
        \prod\limits_{\theta \in \Theta}
        1 - \alpha_{\theta}
    \\
    \label{eq:optimistic-set}
    {}^{\V{\alpha}}_{o} \mm(A)
    &= \mm(A) \cdot
        \prod\limits_{ \substack{\theta \in \Theta \\ A \subseteq \theta} }
        1 - \alpha_{\theta}
        \qquad \forall A \subsetneq \Omega, A \neq \emptyset
    \\
    \label{eq:optimistic-omega}
    {}^{\V{\alpha}}_{o} \mm(\Omega)
    &= \mm(\Omega) \cdot
        \prod\limits_{ \substack{\theta \in \Theta \\ \Omega \subseteq \theta} }
        1 - \alpha_{\theta}
            \\&\nonumber
        + \mm(\emptyset) \cdot
        \prod\limits_{\theta \in \Theta}
        \alpha_{\theta}
            \\&\nonumber
        + \sum\limits_{ A \subseteq \Omega }
        \left[
            \mm(A) \cdot
            \prod\limits_{ \substack{\theta \in \Theta \\ A \subseteq \theta} }
            \alpha_{\theta}
        \right]
\end{align}

\subsection{Proportional discounting}
\label{proportional-discounting}

The above proposed schemes represent two extremes of discounting strategies.
Conservative one that demonstrates very cautious or even overcautious behaviour which can be resumed as: in case of doubt, do not exclude any possibilities.
Indeed, discounting all supersets in the same way as the set in question means that one accepts a possibility that mass of a superset (e.g. $AB$) corresponds \emph{entirely} to one of its constituents (e.g. $A$), which, incidentally, has been overestimated and should hence be discounted.
Conversely, when one assumes that mass of superset $AB$ depends on a set that has not been excessively evaluated ($B$), optimistic discounting is used.
Such a behaviour can be seen as optimistic or bold, because any doubt about whether to discount a particular set or not implies a negative answer.

Since the above schemes are the extreme cases, a need of an in-between solution appears naturally.
A manner of performing this without recurring to mass-dependent computation is to ponder the discount rate by some measure of dependence between a set and it supersets.
The straightforward one is the inclusion criterion measuring the ratio between cardinalities of the set and the superset.
On the basis of this idea, proportional discounting is expressed by:
%
%
\begin{align}
    {}^{\V{\alpha}}_{p} \mm(\emptyset)
        &= \mm(\emptyset) \cdot
        \prod\limits_{\theta \in \Theta}
            1 - \alpha_{\theta}
    \\
    {}^{\V{\alpha}}_{p} \mm(A)
        &= \mm(A) \cdot
        \prod\limits_{ \substack{\theta \in \Theta \\ A \cap \theta \neq \emptyset} }
            1 - \alpha_{\theta}
            \cdot \frac{ \left| A \cap \theta \right| }{ \left| A \right| }
        \qquad \forall A \subsetneq \Omega, A \neq \emptyset
    \\
    {}^{\V{\alpha}}_{p} \mm(\Omega)
        &= \mm(\Omega) \cdot
            \prod\limits_{ \theta \in \Theta }
                1 - \alpha_{\theta}
                \cdot \frac{ \left| \Omega \cap \theta \right| }{ \left| \Omega \right| }
            \\&\nonumber
        + \mm(\emptyset) \cdot
        \prod\limits_{\theta \in \Theta}
        \alpha_{\theta}
            \\&\nonumber
        + \sum\limits_{ A \subseteq \Omega }
            \left[
                \mm(A) \cdot
                \prod\limits_{ \substack{\theta \in \Theta \\ A \cap \theta \neq \emptyset} }
                \alpha_{\theta}
                \cdot \frac{ \left| A \cap \theta \right| }{ \left| A \right| }
            \right]
\end{align}

\section{Properties}
\label{sec:behaviour}
\label{sec:properties}


\subsection{Generalisation of classical discounting}

Proposed discounting schemes generalise classical discounting in the case where \( \Theta = \{ \Omega \} \).
Such a behaviour comes simply from the fact that for any $\theta \in \Theta$, all its subsets will get discounted.
Since all sets are subsets of $\Omega$, all of them are affected in the same way (except for $\Omega$ itself as expected).
%

\subsection{Order invariance}

The result of the discounting operations over different classes is invariant to the order of these operations, equally for conservative, optimistic and for proportional discounting.
The proof is omitted here, as it is trivial and is based on the commutative property of the multiplication.
\begin{align}
    {}_{\Theta_2}^{\V{\alpha}} ({}_{\Theta_1}^{\V{\alpha}} \mm)
    &= {}_{\Theta_1}^{\V{\alpha}} ({}_{\Theta_2}^{\V{\alpha}} \mm)
\end{align}

\subsection{Operation grouping}

\renewcommand{\arraystretch}{1.4} 
\begin{table*}[tb!]
    \centering
    \begin{tabular}{|c||c|c|c|}
        \hline
        \vphantom{\( \sum_{}\limits^{} \)}
        A
        & \( {}_{o}^{\V{\alpha}} \mm(A) \)
        & \( {}_{p}^{\V{\alpha}} \mm(A) \)
        & \( {}_{c}^{\V{\alpha}} \mm(A) \) \\ 
        \hline
        \hline \( \emptyset \)
        & \( \beta_{1} \beta_{2,3} \mm(\emptyset) \)
        & \( \beta_{1} \beta_{2,3} \mm(\emptyset) \)
        & \( \beta_{1} \beta_{2,3} \mm(\emptyset) \) \\
        \hline \( \{ \omega_{1} \} \)
        & \( \beta_{1} \mm(\{ \omega_{1} \}) \)
        & \( \beta_{1} \mm(\{ \omega_{1} \}) \)
        & \( \beta_{1} \mm(\{ \omega_{1} \}) \) \\
        \hline \( \{ \omega_{2} \} \)
        & \( \beta_{2,3} \mm(\{ \omega_{2} \}) \)
        & \( \beta_{2,3} \mm(\{ \omega_{2} \}) \)
        & \( \beta_{2,3} \mm(\{ \omega_{2} \}) \) \\
        \hline \( \{ \omega_{1}, \omega_{2} \} \)
        & \( \mm(\{ \omega_{1}, \omega_{2} \}) \)
        & \( (1 - \frac{1}{2} \cdot \alpha_{1}) (1 - \frac{1}{2} \cdot \alpha_{2,3}) \mm(\{ \omega_{1}, \omega_{2} \}) \)
        & \( \beta_{1} \beta_{2,3} \mm(\{ \omega_{1}, \omega_{2} \}) \) \\
        \hline \( \{ \omega_{3} \} \)
        & \( \beta_{2,3} \mm(\{ \omega_{3} \}) \)
        & \( \beta_{2,3} \mm(\{ \omega_{3} \}) \)
        & \( \beta_{2,3} \mm(\{ \omega_{3} \}) \) \\
        \hline \( \{ \omega_{1}, \omega_{3} \} \)
        & \( \mm(\{ \omega_{1}, \omega_{3} \}) \)
        & \( (1 - \frac{1}{2} \cdot \alpha_{1}) (1 - \frac{1}{2} \cdot \alpha_{2,3}) \mm(\{ \omega_{1}, \omega_{3} \}) \)
        & \( \beta_{1} \beta_{2,3} \mm(\{ \omega_{1}, \omega_{3} \}) \) \\
        \hline \( \{ \omega_{2}, \omega_{3} \} \)
        & \( \beta_{2,3} \mm(\{ \omega_{2}, \omega_{3} \}) \)
        & \( \beta_{2,3} \mm(\{ \omega_{2}, \omega_{3} \}) \)
        & \( \beta_{2,3} \mm(\{ \omega_{2}, \omega_{3} \}) \) \\
        %
        %
        %
        \hline
    \end{tabular}
    \caption{Comparative table of the proposed discounting methods. For succinctness, \( \beta_i = 1 - \alpha_i \).
        \\
        Mass attributed to $\Omega$ omitted for clarity, since for all mass functions \( \mm(\Omega) = 1 - \sum_{A \subsetneq \Omega} \mm(A) \).}
    \label{tab:example-simple}
\end{table*}

For all the proposed schemes, the result of two discounting operations on sets $\Theta_{1}$, $\Theta_{2}$ and discount rate vectors $\V{\alpha}_1$, $\V{\alpha}_2$ done one after another is equal to a single discounting operation on combined discount rate vector $\V{\alpha} = \text{concatenate}(\V{\alpha_{1}}, \V{\alpha_{2}})$.
\begin{align}
    {}_{\Theta_1}^{\V{\alpha}_1} ({}_{\Theta_2}^{\V{\alpha}_2} \mm)
    &= {}_{\Theta_1 \cup \Theta_2}^{\V{\alpha}} \mm
    & \text{ if } \Theta_1 \cap \Theta_2 = \emptyset
\end{align}
This property can be easily generalised for any number of discounting operations.
\begin{align}
    {}_{\Theta_K}^{\V{\alpha}_K} \left( \ldots \left( {}_{\Theta_1}^{\V{\alpha}_1} \mm \right) \ldots \right)
    &= {}_{\Theta}^{\V{\alpha}} \mm
\end{align}
given that
\begin{align}
    \label{eq:operation-grouping-secondary}
    \Theta &= \bigcup\limits_{i \in \{ 1, \ldots, K \} } \Theta_i
    \\
    \V{\alpha} &= \text{concatenate}(\V{\alpha}_1, \ldots, \V{\alpha}_K)
\end{align}
and under the following condition:
\begin{align}
    \label{eq:operation-grouping-condition}
    \forall i, j \in \{ 1, \ldots, K \}, i \neq j &: \Theta_i \cap \Theta_j = \emptyset
\end{align}

\section{Examples}
\label{sec:examples}

\subsection{Example \theExampleCounter: comparison}
\addtocounter{ExampleCounter}{1}
\label{example-simple}

Let \( \Omega = \left\{ \omega_{1}, \omega_{2}, \omega_{3} \right\} \) and let $\mm$ be a \ac{BBA} defined on $\Omega$.
Table~\ref{tab:example-simple} presents the result which yield the proposed discounting schemes with
\( \Theta = \left\{ \left\{ \omega_{1} \right\}, \left\{ \omega_{2}, \omega_{3} \right\} \right\} \)
and discount rate vector \( \V{\alpha} = \left[ \alpha_{1}, \alpha_{2,3} \right] \)%
\footnote{The fact that $\Theta$ represents a partition of $\Omega$ is insignificant, since it could be any subset of $2^{\Omega}$.}%
.
For clarity, we use \( \beta_i = 1 - \alpha_i \).
It is noteworthy that we can arrange the proposed discounting operations in incrementing order of total discounted mass: optimistic $\preceq$ proportional $\preceq$ conservative.
For all mass functions and all discount rate vectors, the following equation holds:
\begin{align}
    \label{eq:discount-order}
    {}_{o}^{\V{\alpha}} \mm(A)
    \geq
    {}_{p}^{\V{\alpha}} \mm(A)
    \geq
    {}_{c}^{\V{\alpha}} \mm(A)
    &
    & \forall A \subsetneq \Omega
\end{align}

\subsection{Example \theExampleCounter: source reliability modelling}
\addtocounter{ExampleCounter}{1}
\label{example-source-reliability}

Let us consider an example of a simplified aerial target recognition problem borrowed from \cite{Elouedi2004, Mercier2006}.
The \acl{FOD} \( \Omega = \{ a,\ h,\ r \} \) contains three classes: air-plane ($a = \omega_{1}$), helicopter ($h = \omega_{2}$) and rocket ($r = \omega_{3}$).
Sensor $S$ provides us with a~\ac{BBA}~$\mm$ hesitating between classifying the target as an air-plane or a rocket:
\begin{align}
    \label{eq:example-targets-initial}
    \mm(\{ a \}) = 0.5
    \qquad
    \mm(\{ r \}) = 0.5
\end{align}

Let us now consider that the sensor is over-reliable when the source is a helicopter or a rocket with plausibility $\alpha_{2,3} = \alpha_{h, r} = 0.4$, while being reliable when the target is an air-plane.
The conservatively discounted \ac{BBA} \( {}_{c}^{\V{\alpha}} \mm \) is:
\begin{align}
    \label{eq:example-targets-discounted-conservative}
    {}_{c}^{\V{\alpha}} \mm(\{ a \}) = 0.5
    \qquad
    {}_{c}^{\V{\alpha}} \mm(\{ r \}) = 0.3
    \qquad
    {}_{c}^{\V{\alpha}} \mm(\Omega) = 0.2
\end{align}
It is to remark that a fraction $0.4$ of the mass attributed to $\{r\}$ has been transferred to $\Omega$, which can be interpreted as follows:
if the target is a helicopter or a rocket, then the source is over-reliable and it might have quantified excessively its belief about target being a helicopter, a rocket or any of the two.
Thus, the target reported as a rocket may in reality be of another type.
%

For completeness, the optimistically discounted \ac{BBA} \( {}_{o}^{\V{\alpha}} \mm \) and the proportionally discounted \ac{BBA} \( {}_{p}^{\V{\alpha}} \mm \) are:
\begin{align}
    \label{eq:example-targets-discounted-optimistic}
    {}_{o}^{\V{\alpha}} \mm(\{ a \}) = 0.5
    \qquad
    {}_{o}^{\V{\alpha}} \mm(\{ r \}) = 0.5
\end{align}
\begin{align}
    \label{eq:example-targets-discounted-proportional}
    {}_{p}^{\V{\alpha}} \mm(\{ a \}) = 0.5
    \qquad
    {}_{p}^{\V{\alpha}} \mm(\{ r \}) = 0.3
    \qquad
    {}_{p}^{\V{\alpha}} \mm(\Omega) = 0.2
\end{align}

A similar example, with \( \alpha_{1} = \alpha_{a} = 0.4 \), using contextual discounting cited from Mercier \cite[Example~2, Case~1]{Mercier2006} gives:
\begin{align}
    \label{eq:example-targets-discounted-contextual}
    {}^{\V{\alpha} \cup} \mm(\{ a \}) = 0.5
    \qquad
    {}^{\V{\alpha} \cup} \mm(\{ r \}) = 0.3
    \qquad
    {}^{\V{\alpha} \cup} \mm(\{ a, r \}) = 0.2
\end{align}
This shows that the behaviour is almost inverse to conservative and proportional discounting and different than optimistic discounting.
Namely, the discount factor being set to the same value but attributed to the compliment set, the resulting mass function is identical.

\section{Case study: temporal discounting}
\label{sec:case-study}

\label{example-temporal-discounting}
\label{sec:temporal-discounting}

In this section, an application to \emph{temporal discounting} is studied.
The principal idea behind this discounting is the fact that a piece of information becomes partially obsolete with time.
This can happen because the entity described by this particular information is dynamic, changes or is not observed any more.
It is important to underline that different pieces of information become obsolete at possibly different rates.
This example motivates why there is a need for introducing new contextual discounting schemes and why the existing one is not sufficient.
The first part demonstrates some postulates about temporal discounting itself.
Next, the existing contextual discounting scheme is applied to temporal discounting.
Finally, the application of the proposed methods is demonstrated.

\subsection{Postulates}

%
The below stated postulates imply that the temporal discounting should be subject to \emph{exponential decay}, similarly to the process of radioactive decay described by Ernest Rutherford in early 1900's \cite{Rutherford1903}.
Indeed, we opt for the solution where the information \quotemarks{decays}, i.e. a piece of information becomes gradually obsolete.

In the following paragraphs, $A$ will denote a set, $A \subseteq \Omega$, about the reliability of which an additional piece of knowledge is available.

\begin{align}
\label{eq:poisson-process}
{}^{t} \mm(A)
&= \mm(A) \cdot e^{ \left( -\lambda t \right) }
\\
\label{eq:poisson-process-lambda}
\lambda &= \frac{ \ln 2 }{ t_{1/2} }
\\
\label{eq:poisson-process-lambda-general}
\lambda &= \frac{ \ln N }{ t_{1/N} }
\end{align}

\subsubsection{Half-life time \texorpdfstring{\( t_{1/2} \)}{t 1/2}}
\label{postulate-half-life}
The mass attributed to a piece of information is two times smaller than the initial mass after half-life time \( t_{1/2} \).
Thanks to this postulate, one can compare the persistence of different information types by comparing their half-life times.
As far as different information persistence measures are considered, it is noteworthy that choosing \quotemarks{life expectancy} (mean time after which a piece of information becomes completely irrelevant) would prohibit the use of exponential functions and so entail some complications.
\begin{align}
    \label{eq:half-life}
    {}^{t_{1/2}} \mm(A)
    &= \frac{ \mm(A) }{ 2 }
\end{align}
More generally, \emph{$1/N$-life time \( t_{1/N} \):}
the mass attributed to a piece of information represents one-$N$th of the initial mass after time \( t_{1/N} \).
\begin{align}
    \label{eq:half-life-general}
    {}^{t_{1/N}} \mm(A)
    &= \frac{ \mm(A) }{ N }
\end{align}
%

\subsubsection{Order invariance}
\label{postulate-order-invariance}

The result of discounting is independent of the order of operations.
\begin{align}
    {}^{t_{2}} \left( {}^{t_{1}} \mm(A) \right)
    &= {}^{t_{1}} \left( {}^{t_{2}} \mm(A) \right)
\end{align}
%

\subsubsection{Only age-dependent}
\label{postulate-age-dependence}

The discounted mass value depends only on the age of the information
and does not on the number of discounting operations.
Indeed, it is desirable that the frequency at which a piece of information gets discounted, does not change the final result.
\begin{align}
    {}^{t_{2}}  \left( {}^{t_{1}} \mm(A) \right)
    &= {}^{t_{1} + t_{2}} \mm(A)
\end{align}
%

\subsection{Temporal discounting using contextual discounting}
\label{sec:contextual-temporal-discounting}

This section will present an attempt to use contextual discounting as presented by Mercier \cite{Mercier2005, Mercier2006} and a counter-example demonstrating that this discounting scheme is not adapted for this aim.

\subsubsection{\texorpdfstring{$\Theta$}{Theta}-discounting}

As presented in Section~\ref{contextual-discounting},
having defined partition \( \Theta \) of the \acl{FOD} \( \Omega \) and discounting rate vector \( \V{\alpha} \) for all elements of \( \Theta \),
%
discounted mass function ${}^{\V{\alpha}}_{\Theta} \mm$ is computed as follows:
\begin{align}
    {}^{\V{\alpha}}_{\Theta} \mm
        &= \mm \opdisj \mm_{\Theta}
\end{align}

\subsubsection{Direct computation of discounting mass function}

Instead of calculating discounting mass function \( \mm_\Theta \) by applying the disjunctive operator,
one can compute it directly using \cite[Proposition~7]{Mercier2005}
:
\begin{align}
    \label{eq:direct-discounting-mass}
    \mm_\Theta(A)
        &= \prod\limits_{\substack{\theta \in \Theta \\ \theta \subseteq A }} \alpha_{\theta}
        \cdot \prod\limits_{\substack{\theta \in \Theta \\ \theta \not\subseteq A}} \left( 1 - \alpha_{\theta} \right)
\end{align}

\subsubsection{Direct computation of discounted mass function}

Once again, direct computation is possible to obtain discounted mass function \( {}^{\V{\alpha}}_{\Theta} \mm \) using the results from Equations~\ref{eq:contextual-discounting} and~\ref{eq:direct-discounting-mass}, which yields%
\footnote{It is supposed that no discount rate has been defined for the empty set.}%
:
\begin{align}
    \label{eq:discounted-mass-direct}
    %
    {}^{\V{\alpha}} \mm(A)
    &= \left( \mm \opdisj \mm_{\Theta} \right) (A)
    \\ \nonumber
    &= \sum_{B \cup C = A} \mm(B) \cdot \mm_{\Theta}(C)
    \\ \nonumber
    &= \sum_{B \subseteq A} \left[ \mm(B) \cdot \sum_{C \subseteq B} \mm_{\Theta}(C) \right]
    \\ \nonumber
    &= \sum_{B \subseteq A} \left[ \mm(B) \cdot \bel_{\Theta}(B) \right]
\end{align}


\subsubsection{Simplified computation of a discounted mass function}

Let us suppose that $\mm$ is a normal mass function, which enables us to simplify Equation~\ref{eq:discounted-mass-direct} for singletons to:
\begin{align}
    \label{eq:discounted-mass-direct-singletons}
    %
    {}^{\V{\alpha}} \mm(\{ \theta \})
        &= \sum_{B \cup C = \theta} \mm(B) \cdot \mm_{\Theta}(C)
        \\ \nonumber
        &= \sum_{C \subseteq \theta} \mm(\theta) \cdot \mm_{\Theta}(C)
        \\ \nonumber
        &= \mm(\{ \theta \}) \cdot \sum_{C \subseteq \theta} \mm_{\Theta}(C)
        \\ \nonumber
        &= \mm(\{ \theta \}) \cdot \bel_{\Theta}(\{ \theta \})
\end{align}

\begin{table*}[htb!]
    \centering
    \begin{tabular}{|c||c|c|c|c|}
        \hline
        \vphantom{\( \sum_{}\limits^{} \)}
        A
        & \( \mm(A) \)
        & \( {}_{o}^{\V{\alpha}} \mm(A) \)
        & \( {}_{p}^{\V{\alpha}} \mm(A) \)
        & \( {}_{c}^{\V{\alpha}} \mm(A) \) \\ 
        \hline
        \hline \( \emptyset \)
        & 0
        & 0
        & 0
        & 0 \\
        \hline \( \{ \omega_{1} \} \)
        & 0.3
        & \( 0.28125 = 0.3 \cdot 0.9375 \)
        & \( 0.28125 = 0.3 \cdot 0.9375 \)
        & \( 0.28125 = 0.3 \cdot 0.9375 \) \\
        \hline \( \{ \omega_{2} \} \)
        & 0.2
        & \( 0.1 = 0.2 \cdot 0.5 \)
        & \( 0.1 = 0.2 \cdot 0.5 \)
        & \( 0.1 = 0.2 \cdot 0.5 \) \\
        \hline \( \{ \omega_{1}, \omega_{2} \} \)
        & 0.2
        & \( 0.2 \)
        & \( 0.1453125 = 0.2 \cdot (1 - \frac{1}{2} \cdot 0.0625) \cdot (1 - \frac{1}{2} \cdot 0.5) \)
        & \( 0.09375 = 0.2 \cdot 0.9375 \cdot 0.5 \) \\
        \hline \( \{ \omega_{3} \} \)
        & 0.2
        & \( 0.03376 = 0.2 \cdot 0.1688 \)
        & \( 0.03376 = 0.2 \cdot 0.1688 \)
        & \( 0.03376 = 0.2 \cdot 0.1688 \) \\
        \hline \( \{ \omega_{1}, \omega_{3} \} \)
        & 0
        & 0
        & 0
        & 0 \\
        \hline \( \{ \omega_{2}, \omega_{3} \} \)
        & 0
        & 0
        & 0
        & 0 \\
        \hline \( \Omega \)
        & 0.1
        & 0.38499
        & 0.4396775
        & 0.49124 \\
        \hline
    \end{tabular}
    \caption{Temporal discounting using the proposed discount schemes. Case~1.}
    \label{tab:example-temporal-case-1}
\end{table*}

\begin{table*}[htb!]
    \centering
    \begin{tabular}{|c||c|c|c|c|c|}
        \hline
        \vphantom{\( \sum_{}\limits^{} \)}
        A
        & \( \mm(A) \)
        & \( {}_{o}^{\V{\alpha}} \mm(A) \)
        & \( {}_{p}^{\V{\alpha}} \mm(A) \)
        & \( {}_{c}^{\V{\alpha}} \mm(A) \)
        & \( {}^{\V{\alpha} \cup} \mm(A) \) \\ 
        \hline
        \hline \( \emptyset \)
        & 0
        & 0
        & 0
        & 0
        & 0 \\
        \hline \( \{ \omega_{1} \} \)
        & 0.3
        & \( 0.12771 = 0.3 \cdot 0.4257 \)
        & \( 0.12771 = 0.3 \cdot 0.4257 \)
        & \( 0.12771 = 0.3 \cdot 0.4257 \)
        & 0.1723 \\
        \hline \( \{ \omega_{2} \} \)
        & 0.2
        & \( 0.1 = 0.2 \cdot 0.5 \)
        & \( 0.1 = 0.2 \cdot 0.5 \)
        & \( 0.1 = 0.2 \cdot 0.5 \)
        & 0.1 \\
        \hline \( \{ \omega_{1}, \omega_{2} \} \)
        & 0.2
        & \( 0.2 \)
        & \( 0.1069275 = 0.2 \cdot (1 - \frac{1}{2} \cdot 0.5743) \cdot (1 - \frac{1}{2} \cdot 0.5) \)
        & \( 0.04257 = 0.2 \cdot 0.4257 \cdot 0.5 \)
        & 0.1391 \\
        \hline \( \{ \omega_{3} \} \)
        & 0.2
        & \( 0.03376 = 0.2 \cdot 0.1688 \)
        & \( 0.03376 = 0.2 \cdot 0.1688 \)
        & \( 0.03376 = 0.2 \cdot 0.1688 \)
        & 0.1662 \\
        \hline \( \{ \omega_{1}, \omega_{3} \} \)
        & 0
        & 0
        & 0
        & 0
        & 0.15 \\
        \hline \( \{ \omega_{2}, \omega_{3} \} \)
        & 0
        & 0
        & 0
        & 0
        & 0.074 \\
        \hline \( \Omega \)
        & 0.1
        & 0.53853
        & 0.6316025
        & 0.69596
        & 0.1983 \\
        \hline
    \end{tabular}
    \caption{Temporal discounting using the proposed discount schemes. Case~2.\\
        Result of Mercier's contextual discounting in the rightmost column.}
    \label{tab:example-temporal-case-2}
\end{table*}

\subsubsection{Use for temporal discounting}

In order to calculate discount rates $\V{\alpha}$ of contextual discounting from parameters $\V{\kappa}$ of temporal discounting, let us compare side by side temporal discounting (Equation~\ref{eq:poisson-process}) as obtained thanks to the above stated postulates:
\begin{align}
    \label{eq:discount-requirement}
    {}^{\V{\alpha}} \mm(\theta)
    &= \mm(\{ \theta \}) \cdot e^{-\lambda_{\theta} t}
    \\ \nonumber
    &= \mm(\{ \theta \}) \cdot \kappa_{\theta}
        & \forall \theta \in \Theta,
        \ 
        0 < \kappa_{\theta} \leq 1
\end{align}
with the simplified expression of contextually discounted mass (Equation~\ref{eq:discounted-mass-direct-singletons}):
\begin{align}
    \label{eq:discount-result}
    {}^{\alpha} \mm(\{ \theta \})
    &= \mm(\{ \theta \}) \cdot \bel_{\Theta}(\{ \theta \})
\end{align}
which, given that \( m(\{ \theta \}) \neq 0, \forall \theta \in \Theta \), yields:
\begin{align}
    \mm(\{ \theta \}) \cdot \kappa_{\theta}
        & \equiv \mm(\{ \theta \}) \cdot \bel_{\Theta}(\theta)
        & / : \mm(\{ \theta \})
    \\
    \kappa_{\theta}
        & \equiv \bel_{\Theta}(\theta)
        \\
        \label{eq:kappa-alpha-comparison}
    \kappa_{\theta}
        & \equiv \prod\limits_{\substack{B \in \Theta \\ B \not\subseteq \theta}} \left( 1 - \alpha_{B} \right)
\end{align}

Let \( K = |\Omega| = |\Theta| \).
Creating a system of equations for all \( \theta \in \Theta \) using Equation~\ref{eq:kappa-alpha-comparison} issues:
\begin{align}
    \label{eq:system-discount-factors}
    \begin{cases}
        \kappa_{\theta_{1}}
            &= \prod\limits_{\substack{B \in \Theta \\ B \not\subseteq \theta_{1}}} \left( 1 - \alpha_{B} \right)
        \\
        & \vdots
        \\
        \kappa_{\theta_{K}}
            &= \prod\limits_{\substack{B \in \Theta \\ B \not\subseteq \theta_{K}}} \left( 1 - \alpha_{B} \right)
    \end{cases}
\end{align}
and by solving it, one obtains:
\begin{align}
    \label{eq:discount-alpha-solution}
    %
    \alpha_i &=
        1 - \sqrt[K-1]{ \frac{ \prod\limits_{j \neq i} \kappa_{\theta_{j}} }{ \kappa_{\theta_{i}}^{K-2} } }
\end{align}
\Note{by convention \( \prod\limits_{i \in \{ \} } x_{i} = 1 \).}

From Equations~\ref{eq:discount-requirement} and~\ref{eq:discount-alpha-solution}, we obtain:
\begin{align}
    \label{eq:kappa-lambda-relation}
    \kappa_{\theta}(t)
    &= e^{-\lambda_{\theta} t}
    \\
    \label{eq:discount-requirement-temporal}
    \alpha_i(t)
    &= 1 - \sqrt[K-1]{ \frac{ \prod\limits_{j \neq i} e^{-\lambda_{j} t} }{ \left( e^{-\lambda_{i} t} \right) ^{K-2} } }
\end{align}

\subsubsection{Example and counterexample}
\label{contextual-temporal-discounting-example}

Let consider two cases~$C_1$ and~$C_2$ of a sensor~$S$ providing a mass function \( \mm^{\Omega} \) and \( \Omega = \{ \omega_1, \omega_2, \omega_3 \} \).
For each \( \omega \in \Omega \), a half-life time~$t_{1/2}$ is known:
\begin{align}
    \V{t}_{1/2,\,C_1} &= [1, 4, 15] \ \si{\second}
    \\
    \V{t}_{1/2,\,C_2} &= [5, 4, 15] \ \si{\second}
\end{align}
Case~1 can be interpreted as follows.
Additional knowledge about source $S$ is available and it states that classes $\omega_{1}$, $\omega_{2}$ and $\omega_{3}$ become obsolete with different rates.
Namely, $\omega_{1}$ is known to be worth a half of its initial value%
\footnote{The word \emph{value} corresponds to some subjective value of a piece of information from the point of view of the fusion system.}
after \num{1} second, $\omega_{2}$ and $\omega_{3}$ --- after \SI{4}{\second} and \SI{15}{\second} respectively.
Analogical interpretation should be given to Case~2 with the sole difference that the half-life period of class~$\omega_{1}$ is longer and equal to \SI{5}{\second}.

Using Equation~\ref{eq:poisson-process-lambda}, decay parameters $\V{\lambda}$ are computed:
\begin{align}
    \V{\lambda}_{C_1} &\approx {[0.6931, 0.1733, 0.0462]}
    \\
    \V{\lambda}_{C_2} &\approx {[0.1386, 0.1733, 0.0462]}
\end{align}
Then, thanks to Equations~\ref{eq:kappa-lambda-relation} and~\ref{eq:discount-requirement-temporal}, let compute parameters $\V{\kappa}$ and discount factor vector $\V{\alpha}$ for instant \( t = \SI{4}{\second} \):
\begin{align}
    \V{\kappa}_{C_1}(t) &\approx {[0.0625, 0.5000, 0.8312]}
    \\
    \V{\kappa}_{C_2}(t) &\approx {[0.5743, 0.5000, 0.8312]}
\end{align}
\begin{align}
    \V{\alpha}_{C_1} &\approx {[-1.5787, 0.6777, 0.8061]}
    \\
    \V{\alpha}_{C_2} &\approx {[0.1493, 0.0228, 0.4122]}
\end{align}

\subsubsection{Comment}
\label{contextual-temporal-discounting-comment}

The above steps demonstrate that the desired temporal discounting cannot be expressed in terms of contextual discounting as proposed in \cite{Mercier2005}.
Indeed, \( \V{\alpha}_{C_1} \) contains a negative value, which is incompatible with this method and the outcome of such a discounting would not satisfy the condition of a mass function as in Equation~\ref{eq:mass-function}.

\subsection{Temporal discounting using proposed discounting schemes}
\label{temporal-discounting-example}

On the contrary to contextual discounting, the proposed methods are expressive enough to reflect the desired behaviour of temporal discounting.
Let us reuse the same two cases evoked in Section~\ref{contextual-temporal-discounting-example}.
The computation of decay parameters $\V{\lambda}$ and $\V{\kappa}$ is common to both methods.
Moreover, discount rate vector values $\V{\alpha}$ correspond directly to values of $\V{\kappa}$ as shown by:
\begin{align}
    %
    \V{\alpha}_1
    &= [\omega_{1} \mapsto 0.0625, \omega_{2} \mapsto 0.5, \omega_{3} \mapsto 0.8312]
    \\
    \V{\alpha}_2
    &= [\omega_{1} \mapsto 0.5743, \omega_{2} \mapsto 0.5, \omega_{3} \mapsto 0.8312]
\end{align}
Tables~\ref{tab:example-temporal-case-1} and~\ref{tab:example-temporal-case-2} show 3 different discounting methods for the two analysed cases.

\section{Conclusion and perspectives}
\label{sec:conclusion}

\ifdefined\FinalVersion\else
\subsection{Conclusion}
\fi

In this article, we have proposed and defined three types of contextual discounting: conservative, proportional and optimistic.
These methods allow fine-grained modelling of the reliability of the sources.
Moreover, the introduced techniques can be applied to temporal discounting which has been described as well.
It has been demonstrated that the existing contextual discounting introduced by Mercier \cite{Mercier2006} is not strong enough to model temporal discounting.

In addition to the already given applications, the authors consider the use of temporal discounting in the context of intelligent transportation perception.
Various object classes seen by a vehicle should not be forgotten at the same rate.
For instance, information about objects recognised as buildings shall be kept longer than static but possibly mobile objects.
In turn, mobile static objects would persist longer than moving objects.

As a practical advantage, one can mention that for a given discount rate vector, factors by which masses are multiplied to obtain discounted mass function can be precomputed and stored for later use.
The computational complexity of such an algorithm grows linearly with the size of the powerset $2^{\Omega}$ equally for time and space.

\ifdefined\FinalVersion\else
\subsection{Perspectives}
\fi

It would be interesting to automatically or semi-automatically define which type of discounting has to be used in particular situation.
Moreover, a profound study of the properties of the proposed discounting rules seems to be significantly important.
These tasks are left for future research.

\section*{Acknowledgements}
{
This work has been supported by the French Ministry of Defence DGA (Direction G\'{e}n\'{e}rale de l'Armement) with a Ph.D. grant delivered to Marek Kurdej.
}


\bibliographystyle{IEEEtran}
\bibliography{PhD-publications-conferences-fusion2013}

\end{document}